\documentclass[10pt,twocolumn,letterpaper]{article}

\usepackage{cvpr}              
\usepackage{dsfont}
\usepackage{bbm}
\usepackage{amsfonts}
\usepackage{amssymb}
\definecolor{cvprblue}{rgb}{0.21,0.49,0.74}
\usepackage[pagebackref,breaklinks,colorlinks,allcolors=cvprblue]{hyperref}

\def\confName{CVPR}
\def\confYear{2026}

\title{On The Application of Linear Attention in Multimodal Transformers}

\author{Armin Gerami\\
University of Maryland, Department of Computer Science and UMIACS\\
{\tt\small agerami@umd.edu}
\and
Seyedehanita Madani\\
Johns Hopkins University, Department of Electrical and Computer Engineering\\
{\tt\small smadani4@jhu.edu}
\and
\and
Ramani Duraiswami\\
University of Maryland, Department of Computer Science and UMIACS\\
{\tt\small ramanid@umd.edu}
}

\begin{document}
\maketitle
\begin{abstract}
Multimodal Transformers serve as the backbone for state-of-the-art vision-language models, yet their quadratic attention complexity remains a critical barrier to scalability. In this work, we investigate the viability of Linear Attention (LA) as a high-efficiency alternative within multimodal frameworks. By integrating LA, we reduce the computational overhead from quadratic to linear relative to sequence length while preserving competitive performance. We evaluate our approach across ViT-S/16, ViT-B/16, and ViT-L/16 architectures trained on the LAION-400M dataset, with validation focused on ImageNet-21K zero-shot accuracy. Our systematic evaluation demonstrates that Linear Attention not only yields significant computational savings but also adheres to the same scaling laws as standard softmax attention. These findings position Linear Attention as a robust, scalable solution for next-generation multimodal Transformers tasked with processing increasingly large and complex datasets.
\end{abstract}    

\section{Introduction}
\label{sec:intro}

Multimodal transformers have become a key architecture for tasks that require understanding information from multiple modalities, such as vision and language~\cite{ma2022multimodal,gabeur2020multi,xu2023multimodal,weerasinghe2024multimodal,tsai2019multimodal}. Their success largely stems from the self- and cross-attention mechanisms, which enable models to capture both intra-modal and cross-modal relationships. As a result, transformer-based models have achieved strong performance in applications such as visual question answering, image captioning, and multimodal retrieval.

Despite these advances, standard attention mechanisms remain computationally expensive. Self-attention scales quadratically with sequence length, which becomes problematic in multimodal settings where inputs often contain hundreds or thousands of visual tokens in addition to text tokens. This quadratic scaling limits the applicability of multimodal transformers to long sequences and high-resolution inputs.

Linear attention (LA) has been proposed as an efficient alternative to softmax attention by reformulating the attention operation to achieve linear scaling with sequence length~\cite{pmlr-v119-katharopoulos20a,han2023flatten,cai2022efficientvit,arora2024simple,wangLinformerSelfAttentionLinear2020,shen2021efficient,qin2023transnormerllm,qin2024lightning,Qin2022-mp}. Recent work has further improved LA using learnable kernels and gating mechanisms~\cite{schlag2021linear,yang2023gated,pmlr-v162-hua22a,yang2024gated}. While these approaches have shown promise in language and vision tasks, their effectiveness in multimodal representation learning remains less explored.

In this work, we investigate the use of LA in multimodal transformers. Since multimodal inputs often involve long token sequences, the linear scaling of LA provides a natural advantage. We evaluate LA by pretraining ViT-S-16, ViT-B-16, and ViT-L-16 models~\cite{dosovitskiy2020vit} on the LAION-400M dataset~\cite{schuhmann2021laion} using the OpenCLIP framework~\cite{ilharco_gabriel_2021_5143773}. Our contributions are:

\begin{itemize}
    \item We investigate the use of linear attention in multimodal transformers by training ViT-S-16, ViT-B-16, and ViT-L-16 models on LAION-400M and show that LA achieves performance comparable to standard attention.
    \item We analyze the computational efficiency of LA in the multimodal setting.
    \item We show that LA follows similar accuracy-model size scaling laws as standard attention.
\end{itemize}

\section{Background}
\label{sec:back}

We briefly review large multimodal models and the computational limitations of standard attention, then introduce Linear Attention (LA), which serves as the basis of our approach.

\subsection{Large Multimodal Models}

The success of large language models (LLMs)~\cite{thoppilan2022lamda,devlin2019bert,brown2020language} has motivated the development of large multimodal models (LMMs) that jointly process visual and textual inputs~\cite{chen2023pali,chen2022pali,driess2023palm,li2023blip}. A common design pairs a pretrained image encoder with a Transformer-based language model, enabling the system to reason across modalities~\cite{tsimpoukelli2021multimodal}. 

Several architectures follow this paradigm. PaLI~\cite{chen2023pali,chen2022pali} combines Vision Transformer encoders with an encoder--decoder language model, while PaLM-E~\cite{driess2023palm} treats visual encoders as additional sensory inputs to a language model. BLIP-2~\cite{li2023blip} instead introduces a lightweight querying transformer that connects frozen vision and language backbones.

A widely used alternative is Contrastive Language--Image Pre-training (CLIP)~\cite{radford2021learning,ilharco_gabriel_2021_5143773}. CLIP uses two separate transformer-based encoders: a vision encoder that maps images to embeddings and a text encoder that maps captions to text embeddings. Both representations are projected into a shared embedding space and trained with a symmetric contrastive objective that pulls matching image-text pairs together while pushing mismatched pairs apart. This training scheme enables strong zero-shot transfer, allowing the model to classify images by comparing their embeddings with text prompts such as ``a photo of a dog''.
As multimodal transformers scale to larger models and longer contexts, improving their computational efficiency becomes increasingly important. In this work we explore replacing standard attention with LA to reduce the computational cost.\\

\subsection{Standard Attention and the Memory Bottleneck}

For a sequence of length $N$ and head dimension $D$, standard attention computes the output $\mathbf{O}\in\mathbb{R}^{N\times D}$ using the Softmax-normalized attention matrix $\mathbf{A}$:

\begin{equation}
\mathbf{O} = \mathbf{A}\mathbf{V},\quad
\mathbf{A} = \mbox{Softmax}\left(\mathbf{Q}\mathbf{K}^T/\sqrt{D}\right),
\label{eq:softmax_attn}
\end{equation}

\begin{equation}
\mathbf{o}_{ij} =
\dfrac{\sum_{n=1}^{N} \exp(\mathbf{q}_i\cdot \mathbf{k}_n/\sqrt{D})\,\mathbf{v}_{nj}}
{\sum_{n=1}^{N} \exp(\mathbf{q}_i\cdot \mathbf{k}_n/\sqrt{D})},
\label{eq:kernel_view}
\end{equation}

where the exponential function acts as the attention kernel. Computing $\mathbf{A}$ requires $O(N^2D)$ operations and acts as the computational bottleneck. Hardware-aware implementations such as FlashAttention~\cite{dao2022flashattention,dao2024flashattention,shah2024flashattention,dong2024flex} provide a significant speedup through tiling and improved data-movement, but the quadratic time complexity remains that limits scalability for long sequences.

\subsection{Linear Attention}Linear Attention (LA) addresses the quadratic bottleneck by replacing the Softmax kernel with a decomposable feature map $\phi(\cdot)$ such that $f(\mathbf{q}, \mathbf{k}) = \phi(\mathbf{q})^T \phi(\mathbf{k})$~\cite{pmlr-v119-katharopoulos20a}. Exploiting the associativity of matrix multiplication, the computation order can be rewritten:
\begin{equation}\mathbf{O} = \dfrac{(\phi(\mathbf{Q})\phi(\mathbf{K})^T)\mathbf{V}}{\phi(\mathbf{Q})\phi(\mathbf{K})^T\,\mathds{1}} = \dfrac{\phi(\mathbf{Q})(\phi(\mathbf{K})^T\mathbf{V})}{\phi(\mathbf{Q})(\phi(\mathbf{K})^T\,\mathds{1})},\label{eq:3}\end{equation}
where $\mathds{1}$ is all ones and $\mathds{1}\in \mathbb{R^{N\times D}}$. Since $\mathbf{Q},\,\mathbf{K},\,\mathbf{V}\in \mathbb{R^{N\times D}}$, the total computational complexity of Equation~\ref{eq:3} becomes $O(ND^2)$.

\section{Model}
\label{sec:meth}
In this work, we employ an affine kernel defined as $f(x) = 1 + x$. By applying this kernel to Equation~\ref{eq:kernel_view} with (left) and without (right) a causal mask, the formulation becomes:
\begin{equation}
\mathbf{o}_{ij} = \dfrac{\sum_{n=1}^{i} (1+\mathbf{q}_i.\mathbf{k}_n)\,\mathbf{v}_{nj}}{\sum_{n=1}^{i} 1+\mathbf{q}_i.\mathbf{k}_n},\quad = \dfrac{\sum_{n=1}^{N} (1+\mathbf{q}_i.\mathbf{k}_n)\,\mathbf{v}_{nj}}{\sum_{n=1}^{N} 1+\mathbf{q}_i.\mathbf{k}_n},\label{eq:4} \end{equation}
which can be computed in $O(ND^2)$ time~\cite{pmlr-v119-katharopoulos20a, gerami2024fast}. Here, $\mathbf{q}_i \cdot \mathbf{k}_n$ represents the raw attention score. Unlike the exponential kernel, the affine kernel may yield negative scores if $\mathbf{q}_i \cdot \mathbf{k}_n < -1$, leading to undefined or counterintuitive behavior. To mitigate this, we normalize the query and key vectors such that:
\begin{gather}
    \mathbf{q}_i = \dfrac{\mathbf{q}_i}{||\mathbf{q}_i||}, \quad \mathbf{k}_i = \dfrac{\mathbf{k}_i}{||\mathbf{k}_i||}, \quad \forall 1 \leq i \leq N,
\end{gather}
This ensures that $-1 \leq \mathbf{q}_i \cdot \mathbf{k}_n \leq 1$, thereby bounding the kernel output to non-negative values.\par
\begin{figure}[h]
    \centering
    \includegraphics[width=1.0\linewidth]{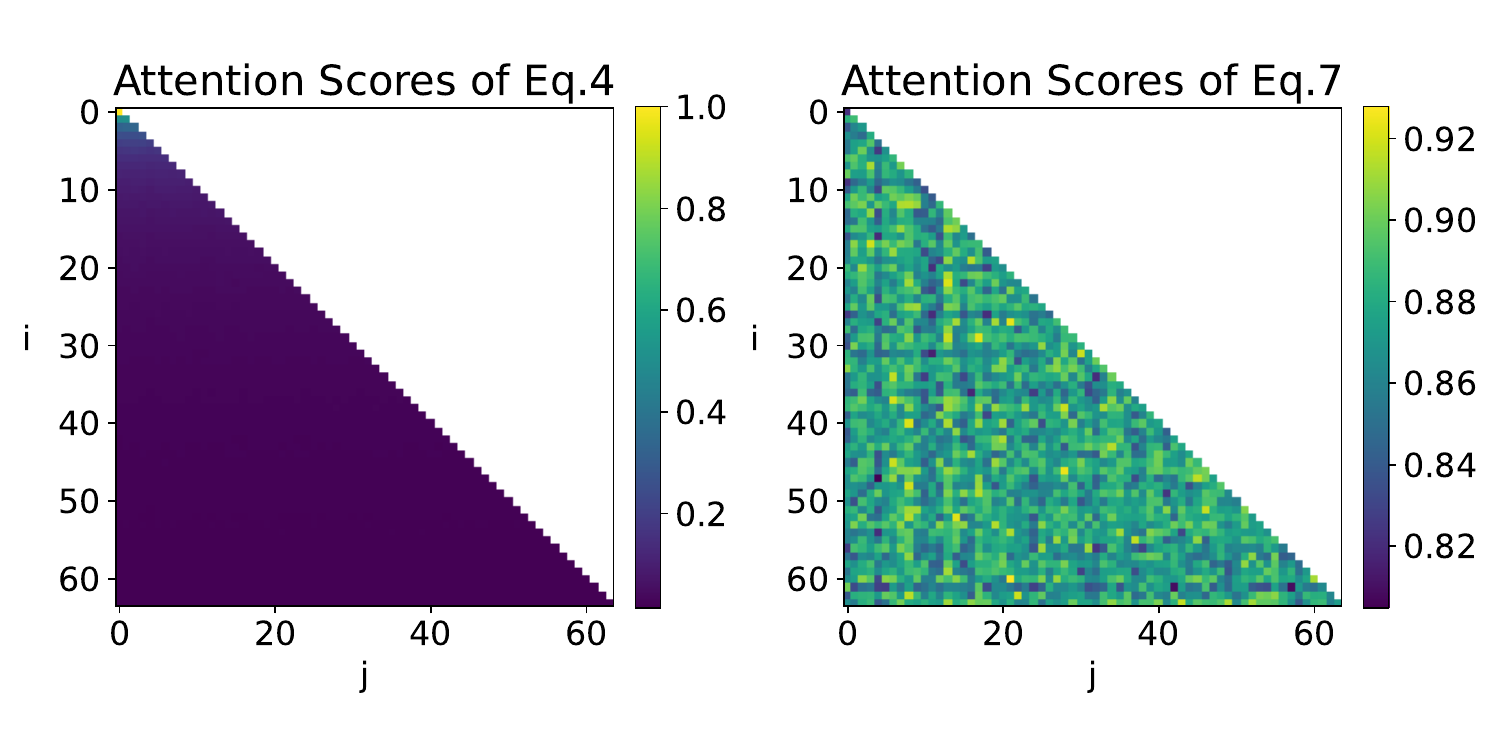}
    \caption{Heatmap visualization of attention scores $a_{ij}$ of Eq.~\ref{eq:4} and \ref{eq:40}. As $i$ increases, the attention scores of Eq.~\ref{eq:4} smooth-out, making it difficult for the model to distinguish relevant query and key pairs. The query and key are random matrices }
    \label{fig:att}
\end{figure}

 While this normalization addresses the issue of negative scores, it imposes an upper bound on the attention scores that decreases with sequence length:
\begin{gather}0 \leq \dfrac{1+\mathbf{q}_i \cdot \mathbf{k}_n}{\sum_{n=1}^{i} (1+\mathbf{q}_i \cdot \mathbf{k}_m)} \leq \dfrac{2}{i}.
\end{gather}
To address this decay, we propose omitting the denominator in Equation~\ref{eq:4}. In standard attention mechanisms, this division serves as a normalizer to ensure weights sum to one. However, by pre-normalizing $\mathbf{Q}$ and $\mathbf{K}$, the individual terms $(1+\mathbf{q}_i \cdot \mathbf{k}_n)$ are already naturally bounded. Therefore, we replace Equation~\ref{eq:4} with:
\begin{gather}
\mathbf{o}_{ij} = \dfrac{\sum_{n=1}^{i} (1+\mathbf{q}_i \cdot \mathbf{k}_n)\,\mathbf{v}_{nj}}{2}, \label{eq:40}\end{gather}
which maintains attention scores between 0 and 1. To visualize the impact of this modification, consider Figure~\ref{fig:att}. We generated normalized random matrices for $\mathbf{Q}$ and $\mathbf{K}$ and computed the resulting attention matrices using Equation~\ref{eq:4} (left) and Equation~\ref{eq:40} (right). As $i$ increases, the attention scores in the left matrix become increasingly uniform (smoother). This over-smoothing makes it difficult for the model to assign high saliency to relevant token pairs, thereby hindering the gradient flow necessary to update model weights effectively during training and limiting the models expresivity. This is in agreement with our results in Section~\ref{sec:4.2}, where the attention from Equation~\ref{eq:4} struggles with the learning task, while our proposed change resolves this issue.

\section{Results}
\label{sec:exp}

We evaluate LA in three experiments: runtime scaling (Sec.~\ref{sec:4.1}), training performance on ViT-S/B/L models (Sec.~\ref{sec:4.2}), and scaling behavior compared to standard attention (Sec.~\ref{sec:4.3}).

\begin{figure}[h]
    \centering
    \includegraphics[width=0.9\linewidth, trim=0mm 115mm 175mm 0mm, clip=true]{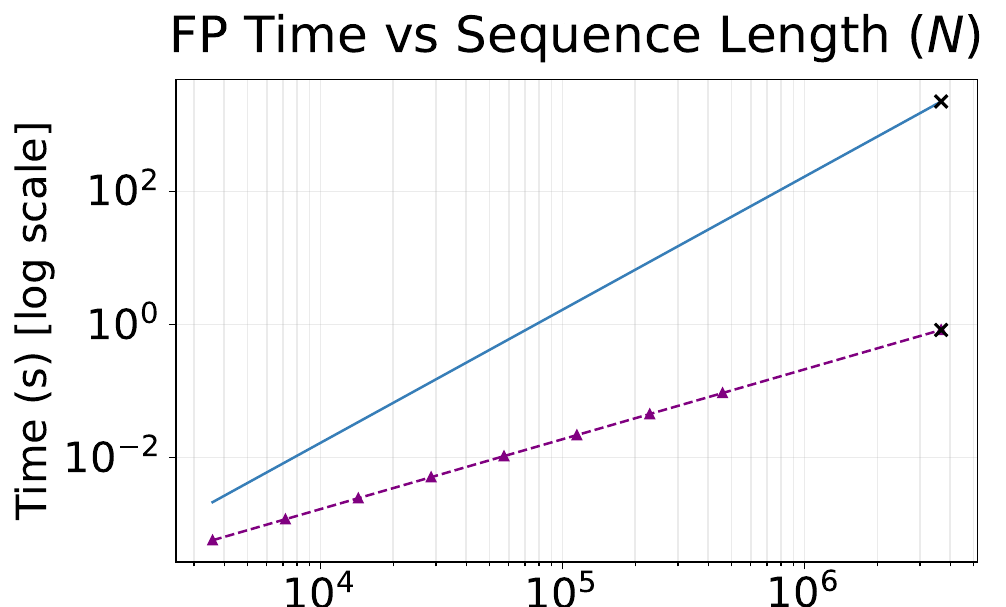}
    \includegraphics[width=0.9\linewidth, trim=173mm 105mm 5mm 0mm, clip=true]{figures/H200.pdf}
    \caption{Forward and backward pass time for a single attention layer with 8 heads, 64 dimensions per head, and batch size of 4 on an H200 GPU.}
    \label{fig:time}
\end{figure}

\subsection{Time Scaling}
\label{sec:4.1}

Standard attention scales as $O(N^2D)$, while LA scales as $O(ND^2)$, making LA increasingly efficient for long sequences. Figure~\ref{fig:time} shows the runtime of a single attention layer with 64 dimensions per head (matching the models in Section~\ref{sec:4.2}) and 8 heads with a batch size 4. Measurements are averaged over 1000 runs after 100 warm-up iterations on an H200 GPU. We use FlashAttention-2~\cite{dao2024flashattention} for standard attention and an optimized LA implementation from~\cite{gerami2025transformer}. The log-log plot slopes show a scaling of $O(N)$ for LA and $O(N^2)$ for standard attention. This improved efficiency is noticeable from token lengths as small as $10^3$, and reaches $\sim 10^3$ lower time in token lengths of $4\times 10^6$.

\begin{figure*}[h]
    \centering
    \includegraphics[width=1.0\textwidth, trim=0mm 0mm 0mm 0mm, clip=true]{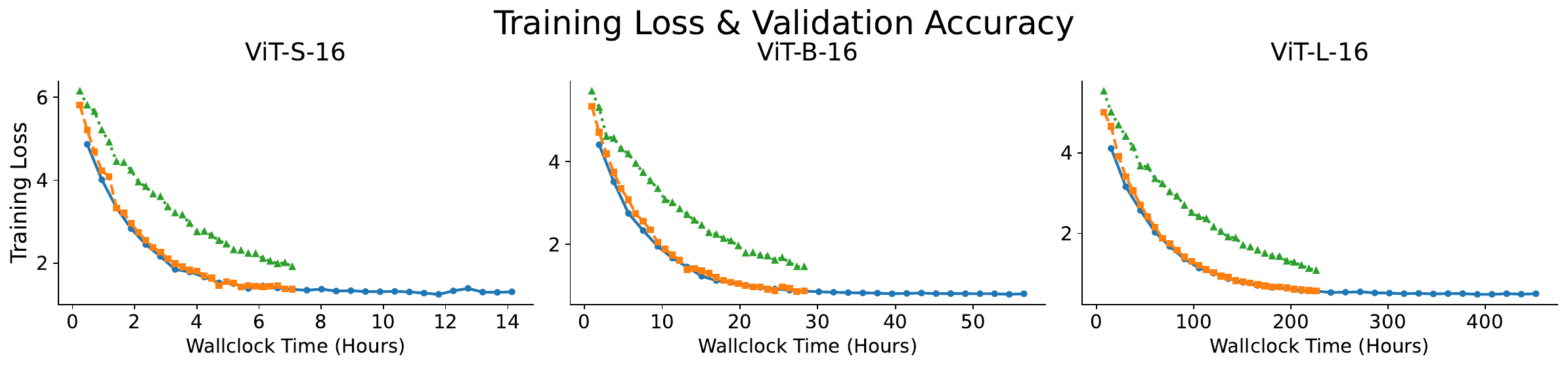}
    \includegraphics[width=1.0\textwidth, trim=0mm 0mm 0mm 23mm, clip=true]{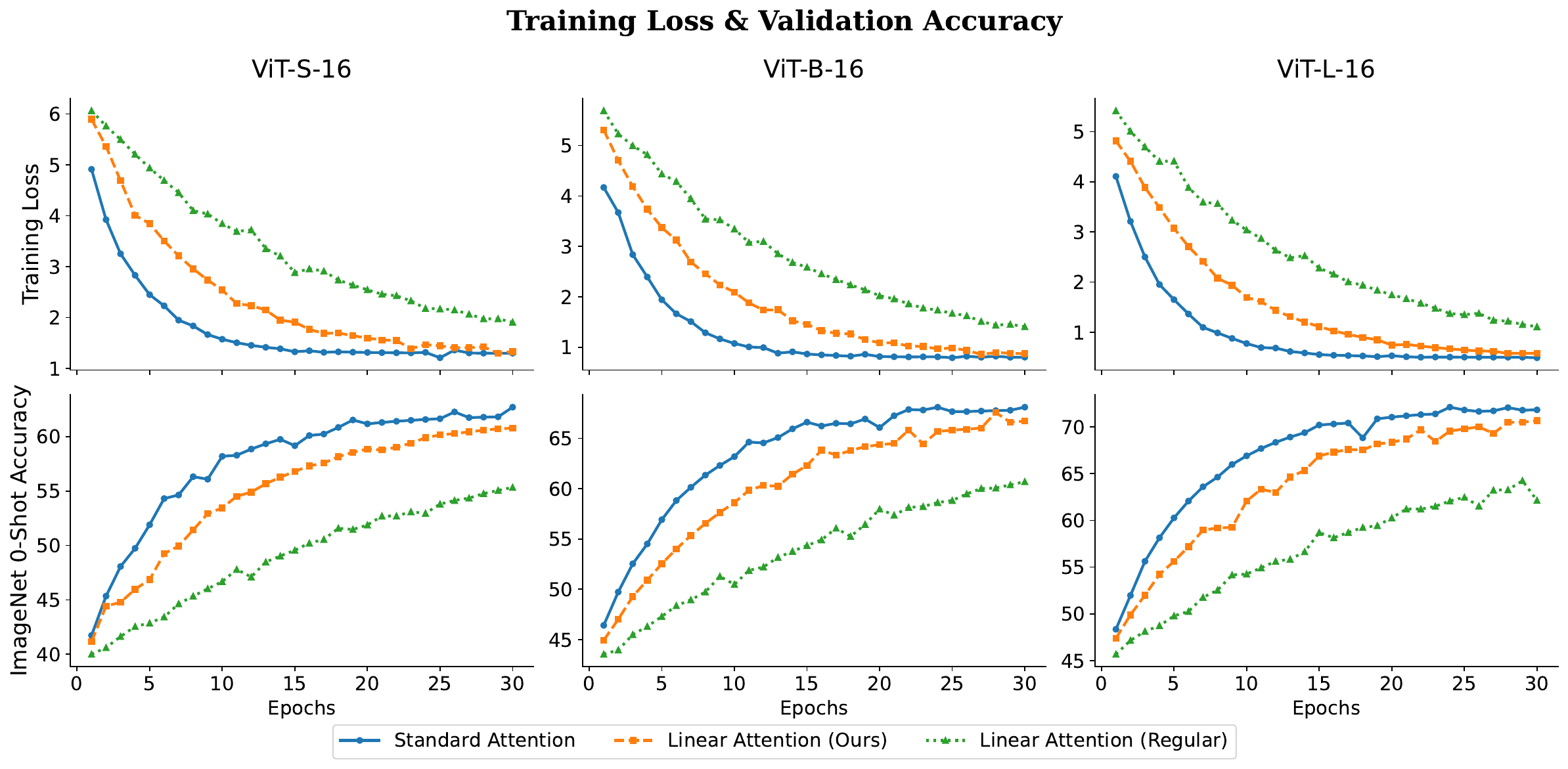}
    \caption{Training metrics for ViT-S/16, ViT-B/16, and ViT-L/16 models trained on LAION-400M using four NVIDIA A5500 GPUs. Linear attention with our adjustments achieves performance comparable to standard attention.}
    \label{fig:training}
\end{figure*}

\subsection{Training Curve}
\label{sec:4.2}

We implement our experiments using OpenCLIP~\cite{ilharco_gabriel_2021_5143773} and train ViT-S-16, ViT-B-16, and ViT-L-16 on LAION-400M~\cite{schuhmann2021laion}. Global batch sizes are 64, 16, and 4 respectively, where the training is conducted on four A5500 GPUs. Model specifications are summarized in Table~\ref{tab:spec}. Validation performance is measured using ImageNet21K zero-shot accuracy~\cite{ridnik2021imagenet21k}.

\begin{table}[h]
    \centering
    \begin{tabular}{l|lccc}
        \toprule
        Model& Param&Layers & Width & Heads \\
        \midrule
        ViT-S-16& 22M&12&384&6 \\
        ViT-B-16& 86M&12&512&8 \\
        ViT-L-16& 304M&24&768&12 \\
        \bottomrule
    \end{tabular}
    \caption{Specifications of the models used in our study.}
    \label{tab:spec}
\end{table}

Figure~\ref{fig:training} illustrates the training trajectories and per-epoch validation accuracy for both methods. From these results, we derive two primary insights. First, LA converges to approximately the same terminal value as standard attention, suggesting that the expressivity of LA is comparable to standard attention in multimodal contexts. However, LA requires more epochs to reach this point. This disparity is likely caused by the linear kernel's limited ability to produce sharp contrast between attention scores. In contrast, the exponential kernel in standard attention generates higher variance in attention weights, allowing the model to more easily identify and prioritize meaningful query-key pairs.
\par Second, the conventional formulation of LA (Equation~\ref{eq:4}) exhibits remarkably slow convergence, a bottleneck that our proposed adjustments (Equation~\ref{eq:40}) significantly mitigate. This improvement is similarly tied to the attention mechanism's capacity for contrast. As demonstrated in Figure~\ref{fig:att}, standard LA produces overly smooth attention maps, whereas our adjustments mitigates this limited attention weight contrast and improves the model's expressivity.

\subsection{Scaling Law}
\label{sec:4.3}
\begin{figure}[b]
    \centering
    \includegraphics[width=1.0\linewidth]{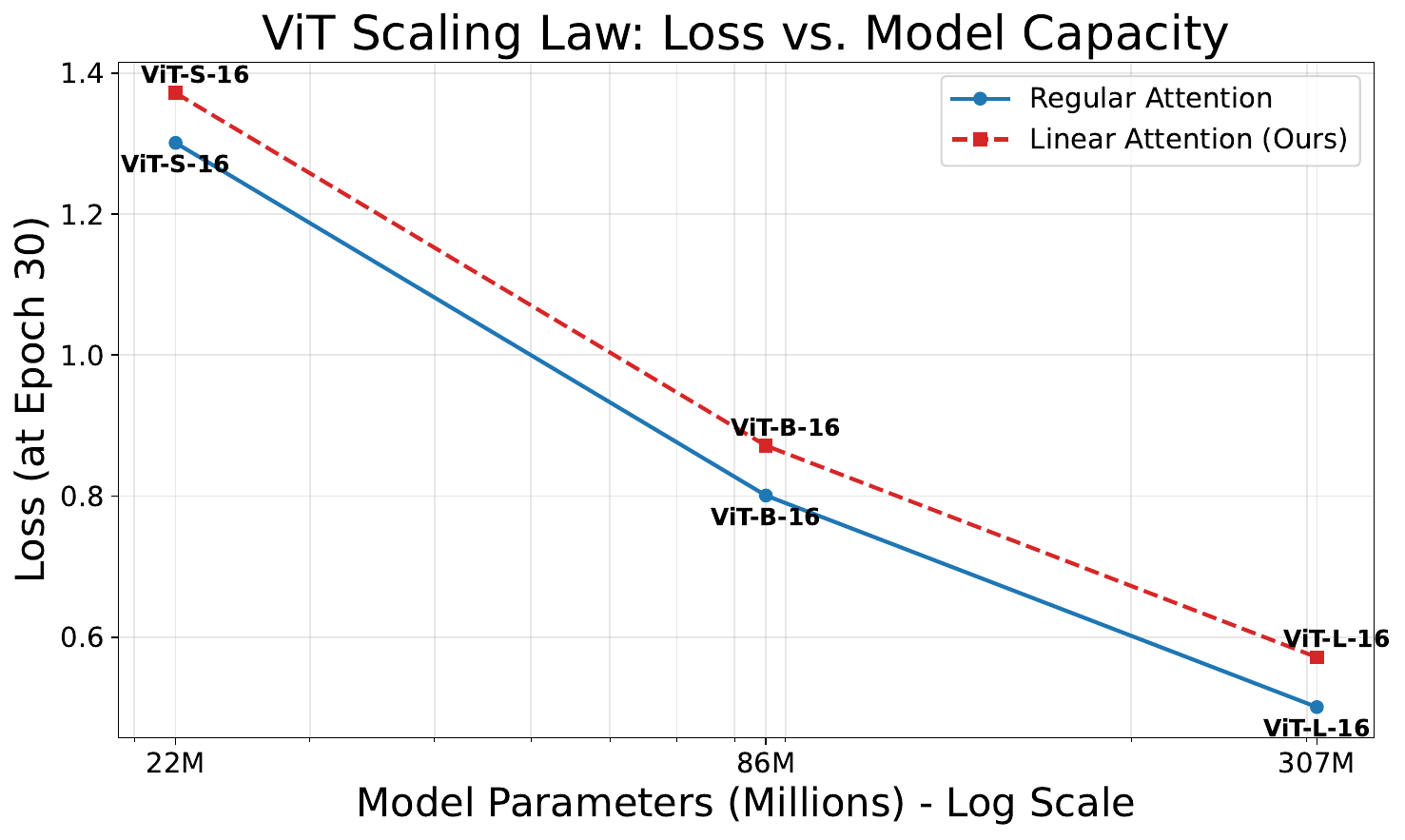}
    \caption{Training loss scaling with model size for ViTs on LAION-400M. Our adjusted LA follows a trend similar to standard attention.}
    \label{fig:scale}
\end{figure}

Large models often follow power-law scaling between loss $L$ and parameter count $N$ expressed as $L(N)\approx aN^{-\alpha}$ that has been observed in language~\cite{kaplan2020scaling} and multimodal~\cite{li2023inverse} models. Figure~\ref{fig:scale} shows that ViT models using our modified LA ($L_{\text{LA}}$) follow scaling trends similar to standard attention ($L_{\text{SA}}$). The empirical fits are:

\begin{gather}
    L_{\text{SA}}(N) \approx 593\, N^{-0.362}, \quad
    L_{\text{LA}}(N) \approx 376\, N^{-0.332}.
\end{gather}

To elaborate, training loss decreases with model size according to a similar power-law trend for both mechanisms, indicating that LA avoids the performance degradation sometimes associated with linearized attention at scale. This suggests that LA can maintain competitive performance while improving computational efficiency regardless of model scale.

\section{Conclusion}
We studied the use of linear attention (LA) in multimodal Vision Transformer models, and proposed adjustments to improve its expresivity. Our results show that LA provides significant computational advantages due to its linear scaling with sequence length. Experiments on ViT-S, ViT-B, and ViT-L models trained on LAION-400M demonstrate that LA can achieve performance comparable to standard softmax attention while offering improved efficiency.

\par Furthermore, Our scaling analysis shows that models using the modified LA follow similar empirical scaling laws to standard attention. These results suggest that linear attention can serve as an efficient alternative to softmax attention for large-scale multimodal transformers, especially in settings with long sequences.
{
    \small
    \bibliographystyle{ieeenat_fullname}
    \bibliography{main}
}


\end{document}